\documentclass[11pt,a4paper]{article}
\usepackage[hyperref]{acl2019}
\usepackage{times}
\usepackage{latexsym}

\usepackage{url}

\usepackage{xcolor}
\usepackage{booktabs,paralist,multirow}
\usepackage{graphicx}
\usepackage{amssymb,amsfonts,amsmath}
\usepackage{bm}

\usepackage[colorinlistoftodos,prependcaption,textsize=scriptsize]{todonotes}

\aclfinalcopy 

\title{An Imitation Learning Approach to Unsupervised Parsing}

\author{Bowen Li$^\dagger$\quad Lili Mou$^\ddagger$\quad Frank Keller$^\dagger$ \\
	$^\dagger$Institute for Language, Cognition and Computation \\
	School of Informatics, University of Edinburgh, UK \\ 
	    $^\ddagger$University of Waterloo, Canada \\
	  \url{bowen.li@ed.ac.uk},\quad \url{doublepower.mou@gmail.com} \\
	  \url{keller@inf.ed.ac.uk} \\
  }

\date{}

\begin{document}
	\maketitle
	
	\setlength{\abovedisplayskip}{5pt}
\setlength{\belowdisplayskip}{5pt}

	\begin{abstract}
	Recently, there has been an increasing interest in
        unsupervised parsers that optimize semantically oriented
        objectives, typically using reinforcement
        learning. Unfortunately, the learned trees often do not match
        actual syntax trees well. \citet{prpn} propose a structured
        attention mechanism for language modeling (PRPN), which
        induces better syntactic structures but relies on ad hoc
        heuristics. Also, their model lacks interpretability as it is
        not grounded in parsing actions. In our work, we propose an
        imitation learning approach to unsupervised parsing, where we
        transfer the syntactic knowledge induced by the PRPN to a
        Tree-LSTM model with discrete parsing actions. Its policy is
        then refined by Gumbel-Softmax training towards a semantically
        oriented objective. We evaluate our approach on the All
        Natural Language Inference dataset and show that it achieves a
        new state of the art in terms of parsing $F$-score,
        outperforming our base models, including the PRPN.\footnote{Our
          code can be found at
          \\\url{https://github.com/libowen2121/Imitation-Learning-for-Unsup-Parsing}}
	\end{abstract}

\section{Introduction}

From a linguistic perspective, a natural language sentence can be thought of as a set of nested constituents in the form of a tree structure \citep{partee2012mathematical}. 
When a parser is trained on labeled treebanks, the predicted constituency trees are useful for various natural language processing (NLP) tasks, including relation extraction \citep{verga2016relation}, text simplification \citep{narayan2014hybrid}, and machine translation \cite{treemt}. However, expensive expert annotations are usually required to create treebanks.

\textit{Unsupervised parsing} (also known as \textit{grammar induction} or \textit{latent tree learning}) aims to learn syntactic structures without access to a treebank during training, with potential uses in low resource or out-of-domain scenarios.
In early approaches, unsupervised parsers were trained by optimizing the marginal likelihood of sentences~\cite{Klein:Manning:04}. 
More recent deep learning approaches \citep{rl-spinn, maillard2017jointly, choi} obtain latent tree structures by reinforcement learning (RL). 
Typically, this involves a secondary task, e.g., a language modeling objective or a semantic task.
However, \citet{isitsyntax} have pointed out that these methods do not yield linguistically plausible structures, and have low self-agreement when randomly initialized multiple times.

Recently, \newcite{prpn} proposed the parsing-reading-predict network (PRPN), which performs language modeling with structured attention. 
The model uses heuristics to induce tree structures from attention scores, and in a replication was found to be the first latent tree model to produce syntactically plausible structures \cite{replication}. 
Structured attention in the PRPN is formalized as differentiable continuous variables, making the model easy to train. 
But a major drawback is that the PRPN does not model tree-building operations directly. 
These operations need to be stipulated externally, in an ad hoc inference procedure which is not part of the model and cannot be trained (see Section~\ref{sec:approach}).

In this paper, we propose an imitation learning framework that combines the continuous PRPN with a Tree-LSTM model with discrete parsing actions, both trained without access to labeled parse trees. We exploit the advantages of the PRPN by transferring its knowledge to a discrete parser which explicitly models tree-building operations. We accomplish the knowledge transfer by training the discrete parser to mimic the behavior of the PRPN. Its policy is then refined using straight-through Gumbel-Softmax~\cite[ST-Gumbel,][]{st-gumbel} trained with a semantic objective, viz., natural language inference~(NLI).

We evaluate our approach on the All Natural Language Inference dataset and show that it achieves a new state of the art in terms of parsing $F$-score, outperforming our base models, including the PRPN. Our work also shows that semantic objectives can improve unsupervised parsing, contrary to earlier claims \citep{isitsyntax,replication}.

\section{Related Work}

Recursive neural networks are a type of neural network which incorporates syntactic structures for sentence-level understanding tasks.
Typically, recursive neural network models assume that an annotated treebank or a pretrained syntactic parser is available \citep{recursive, treelstm2, kim2018dynamic},
but recent work pays more attention to learning syntactic structures in an unsupervised manner.
\citet{rl-spinn} propose to use reinforcement learning, and \citet{maillard2017jointly} introduce the Tree-LSTM to jointly learn sentence embeddings and syntax trees, later combined with a Straight-Through Gumbel-Softmax estimator by \citet{choi}. 
In addition to sentence classification tasks, recent research has focused on unsupervised structure learning for language modeling \citep{prpn, ordered, diora, urnng}. 
In our work, we explore the possibility for combining the merits of both sentence classification and language modeling.

Unsupervised parsing is also related to differentiation through discrete variables, where researchers have proposed to use reinforcement learning with sampling~\cite{reinforce}, neural attention for marginalization~\cite{deng2018latent}, and proximal gradient methods~\cite{st-gumbel,spigot}. Our work follows the framework of \citet{coupling}, who couple neural and symbolic systems for table querying by pretraining an reinforcement learning executor with neural attention. 
We extend this idea to syntactic parsing and show the relationship between parsing and downstream tasks. Such a framework couples diverse models at the intermediate output level (latent trees in our case); its flexibility allows us to make use of heterogeneous models, such as the PRPN and the Tree-LSTM.

The knowledge transfer between the PRPN and the Tree-LSTM applies a simple imitation learning procedure, where an agent learns from a teacher (a human or a well-trained model) based on demonstrations (i.e.,~predictions of the teacher). Typical approaches to imitation learning include behavior cloning (step-by-step supervised learning) and inverse reinforcement learning \citep{imitation}.  If the environment/simulator is available, the agent can refine its policy after learning from demonstrations \citep{imperfect}.
Our work also adopts a two-step strategy: learning from demonstrations and refining policy. 
Policy refinement is needed in our approach because the teacher is imperfect, and experiments show the benefit of policy refinement in our case.

\section{Our Approach}
\label{sec:approach}

\paragraph{Parsing-reading-predict network (PRPN).}
The first ingredient of our approach is the PRPN, which is trained using a language modeling objective, i.e., it predicts the next word in the text, based on previous words.

The PRPN introduces the concept of \textit{syntactic distance} $d_t$, defined as the height of the common ancestor of $w_{t-1}$ and $w_t$ in the tree ($t$ is the position index in a sentence $w_1, ..., w_N$). 
Since gold standard $d_t$ is not available, the PRPN learns the estimated $\widehat d_t$ end-to-end in an unsupervised manner.
The PRPN computes the differences between $\widehat d_t$ at the current step and all previous steps $\widehat d_j$ for $2 \le j < t$. The differences are normalized to $[0,1]$ and used to compute attention scores right to left. These scores are applied to reweight another set of inner-sentence attention scores, which are then used in a recurrent neural network to predict the next word. The PRPN is explained in more detail in Appendix~\ref{sec:prpn detail}.

Based on the real-valued syntactic distances in the PRPN, an external procedure is used to infer tree structures.
The main text of \citet{prpn} suggests using the following intuitive scheme: find the largest distance $\widehat d_i$ and split the sentence into two constituents $(\cdots, w_{i-1})$ and $(w_i, \cdots)$. This process is then repeated recursively on the two new constituents.

The trees inferred by this scheme, however, yield poor parsing $F$-scores, and the results reported by \newcite{prpn} are actually obtained by a different scheme (evidenced in their supplementary material and code repository): find the largest syntactic distance $\widehat d_i$ and obtain two constituents $(\cdots, w_{i-1})$ and $(w_i,\cdots)$. 
If the latter constituent contains two or more words, then it is further split into $(w_i)$ and $(w_{i+1}, \cdots)$, regardless of the syntactic distance $\widehat d_{i+1}$. This scheme introduces a bias for right-branching trees, which presumably is the reason why it yields good parsing $F$-scores for English.

The reliance on this trick illustrates the point we make in the Introduction: syntactic distance has the advantage of being a continuous value, which can be computed as an attention score in a differentiable model. However, this comes at a price: the PRPN does not model trees or tree-building operations directly. These operations need to be stipulated externally in an ad hoc inference procedure. This procedure is not part of the model and cannot be trained, but yet is crucial for good performance.

\begin{figure}[!t]
	\includegraphics[width=\linewidth]{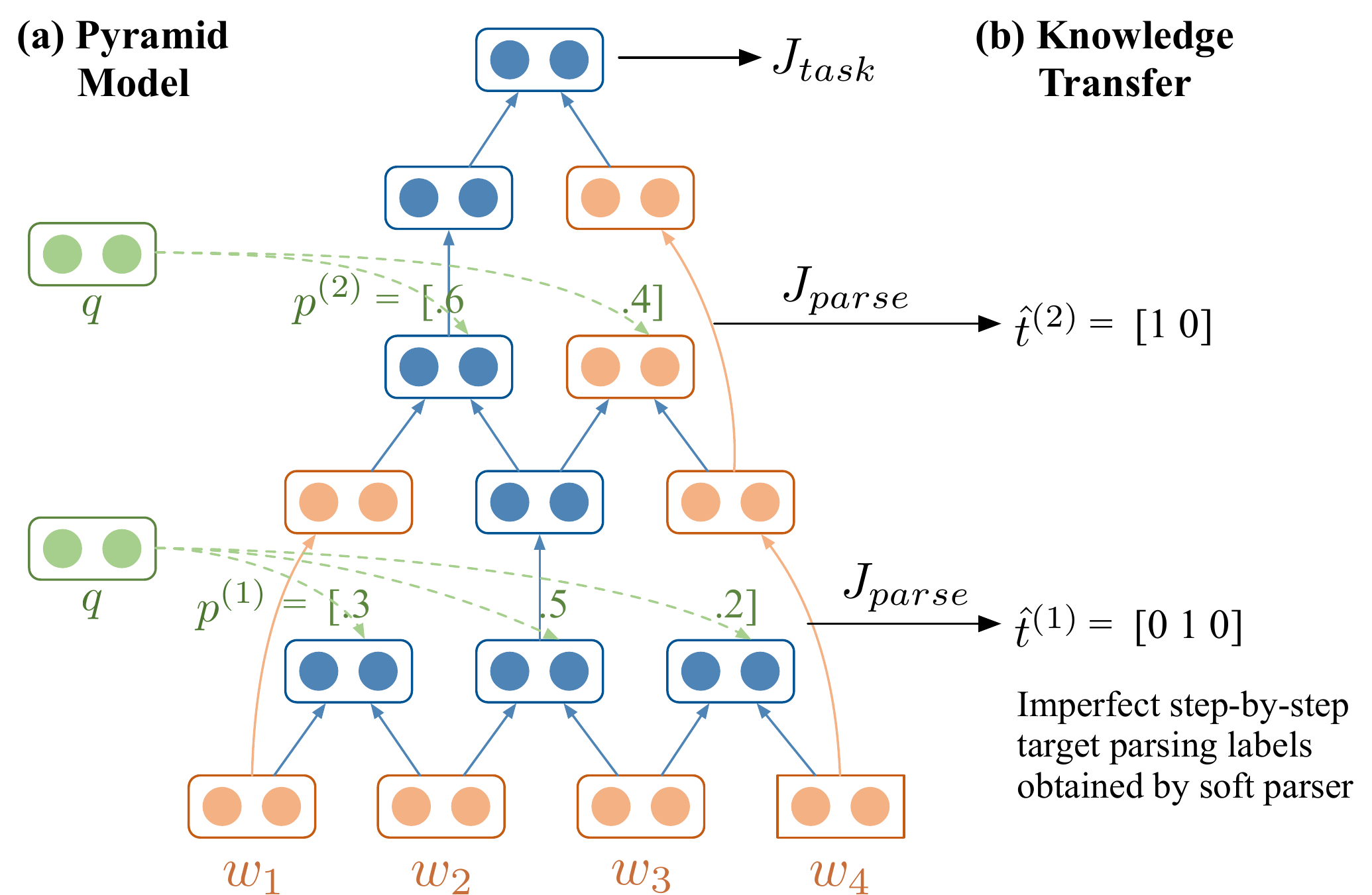}
	\caption{Overview of our approach. (a)~The Tree-LSTM model of \protect\citet{choi}. (b)~The model is first trained with step-by-step supervision, and then Gumbel-Softmax is applied to refine the policy.}\label{fig:overview}
\end{figure}

\paragraph{Discrete syntactic parser.}
To address this problem, we combine the PRPN with a parser which explicitly models tree-building operations. Specifically, we use the pyramid-shaped, tree-based long short-term memory~\cite[Tree-LSTM, Figure~\ref{fig:overview}a,][]{choi}, where reinforcement learning (RL) in this model can be relaxed by Gumbel-Softmax.

Concretely, let $\bm w_1,\bm w_2, \cdots,\bm w_N$ be the embeddings of the words in a sentence. The model tries every possible combination of two consecutive words by the Tree-LSTM, but then uses softmax (in $N-1$ ways) to predict which composition is appropriate at this step. 

Let $\bm h_1^{(1)},\cdots,\bm h_{N-1}^{(1)}$ be the candidate Tree-LSTM composition at the bottom layer. With $\bm q$ being a trainable query vector, the model computes a distribution $\bm p$:
\begin{equation}
p_i^{(1)}=\operatorname{softmax}\{\bm q^\top\bm h_i^{(1)}\}
\end{equation}
Assuming the model selects an appropriate composition at the current step, we copy all other words intactly, shown as orange arrows in Figure~\ref{fig:overview}a. 
This process is applied recursively, forming the structure in the figure.

The Tree-LSTM model is learned by straight-through Gumbel-Softmax (detailed in Appendix~\ref{sec:gumbel detail}), which resembles RL as it samples actions from its predicted probabilities, exploring different regions of the latent space other than a maximum a \textit{posteriori} tree. 
Training involves doubly stochastic gradient descent~\cite{rationale}: the first stochasticity comes from sampling input from the data distribution, and the second one from sampling actions for each input. 
Therefore, ST-Gumbel is difficult to train (similar to RL), and may be stuck in poor local optima, resulting in low self-agreement for multiple random initializations \citep{isitsyntax}.

\paragraph{Imitation learning.} 
Our aim is to combine the PRPN and its continuous notion of syntactic distance with a parser that has discrete tree-building operations. The mapping from the sequence of Tree-LSTM composition operations to a tree structure is not injective. Given a parse tree, we may have multiple different composition sequences, e.g., left-to-right or right-to-left. This ambiguity could confuse the Tree-LSTM during training. We solve this problem by using the PRPN's notion of syntactic distance. 

Given a parse tree predicted by the PRPN, if more than one composition is applicable, we always group the candidates with the lowest syntactic distance.
In this way, we can unambiguously determine the composition order from the trees inferred by the PRPN.
Then, we train the Tree-LSTM model in a \textit{step-by-step}~(SbS) supervised fashion.
Let $\widehat{\bm t}^{(j)}$ be a one-hot vector for the $j$th step of Tree-LSTM composition, where the hat denotes imperfect target labels induced by the PRPN's prediction. The parsing loss is defined as:
\begin{equation}
J_\text{parse}= - \sum\nolimits_j\sum\nolimits_i\widehat t_i^{(j)}\log p_i^{(j)}
\end{equation}
where $\bm p^{(j)}$ is the probability predicted by the Tree-LSTM model. The subscript $i$ indexes the $i$th position among in $1,\cdots,N_j-1$, where $N_j$ is the number of nodes in the $j$th composition step.

The overall training objective $J$ is a weighted combination of the loss of the downstream task and the parsing loss, i.e.,  $J=J_\text{task} + \lambda J_\text{parse}$. After step-by-step training, we perform \textit{policy refinement} by optimizing $J_\text{task}$ with ST-Gumbel, so that the Tree-LSTM can improve its policy based on a semantically oriented task.

It should be emphasized that how the Tree-LSTM model builds the tree structure differs between step-by-step training and ST-Gumbel training. For SbS training, we assume an imperfect parsing tree is in place; hence the Tree-LSTM model exploits existing partial structures to predict the next composition position. For ST-Gumbel, the tree structure is sampled from its predicted probability, enabling our model to explore the space of trees beyond the given imperfect tree.

\section{Experiments}

\begin{table*}
	\label{tab:parsing result}
	\small
	\centering
	\begin{tabular}{l | c c c | c c c} 
		\noalign{\hrule height 1.0pt}    
		\multicolumn{1}{l}{} & \multicolumn{3}{| c |}{ w/o Punctuation } & \multicolumn{3}{c}{w/ Punctuation} \\
		Model                       & Mean $F$      & Self-agreement& RB-agreement  &  Mean $F$     & Self-agreement  &  RB-agreement \\
		\noalign{\hrule height 1.0pt}
		Left-Branching              & 20.7              & -             & -             & 18.9          & -             & -    \\
		Right-Branching             & \textbf{58.5}     & -             & -             & 18.5          & -             & -   \\
		Balanced-Tree               & 39.5              & -             & -             & 22.0          & -             & -   \\
		\hline
		ST-Gumbel                   & 36.4              & 57.0          & 33.8          & 21.9          & 56.8          & \textbf{38.1} \\
		PRPN                        & 46.0              & 48.9          & 51.2          & 51.6          & 65.0          & 27.4   \\
		Imitation (SbS only)        & 45.9              & 49.5          & 62.2          & 52.0          & \textbf{70.8} & 20.6   \\
		Imitation (SbS + refine)    & \,\,53.3$^\dag$   & \textbf{58.2} & \textbf{64.9} & \,\,\textbf{53.7}$^\dag$ & 67.4          & 21.1   \\
		\noalign{\hrule height 1.0pt}
	\end{tabular}
	\caption{Parsing performance with and without punctuation. Mean $F$ indicates mean parsing $F$-score against the Stanford Parser (early stopping by $F$-score). Self-/RB-agreement indicates self-agreement and agreement with the right-branching baseline across multiple runs. $\dag$ indicates a statistical difference from the corresponding PRPN baseline with $p < 0.01$, paired one-tailed bootstrap test.\footnotemark}
	\label{tab:res}
\end{table*}

We train our model on the AllNLI dataset and evaluate on the MultiNLI development set, following experimental settings in \newcite{replication} (for detailed settings, please see Appendix~\ref{setting}).

Table \ref{tab:res} shows the parsing $F$-scores against the Stanford Parser. The ST-Gumbel Tree-LSTM model and the PRPN were run five times with different initializations, each known as a trajectory.
For imitation learning, given a PRPN trajectory, we perform SbS training once and then policy refinement for five runs.
Left-/right-branching and balanced trees are also included as baselines. 

\paragraph{Parsing results with punctuation.} 
It is a common setting to keep all punctuation for evaluation on the AllNLI dataset \citep{replication}. 
In such a setting, we find that the Tree-LSTM, trained by ST-Gumbel from random initialization, does not outperform balanced trees, whereas the PRPN outperforms it by around 30 points. Our PRPN replication results are consistent with \newcite{replication}.
Our first stage in imitation learning (SbS training) is able to successfully transfer the PRPN's knowledge to the Tree-LSTM, achieving an $F$-score of 52.0, which is clearly higher than the 21.9 achieved by the Tree-LSTM trained with ST-Gumbel alone, and even slightly higher than the PRPN itself. 
The second stage, policy refinement, achieves a further improvement in unsupervised parsing, outperforming the PRPN by 2.1 points. 

We also evaluate the self-agreement by computing the mean $F$-score across 25 runs for policy refinement and five runs for other models.
We find that our imitation learning achieves improved self-agreement in addition to improved parsing performance.

\paragraph{Parsing results without punctuation.}
We are interested in investigating whether punctuation make a difference on unsupervised parsing.
In the setting without punctuation, our imitation learning approach with policy refinement outperforms the PRPN by a larger margin (7.3 $F$-score points) than in the setting with punctuation.
But surprisingly, strictly right-branching trees are a very strong baseline in this setting, achieving the best parsing performance overall. The PRPN cannot outperform the right-branching baseline, even though it uses a right-branching bias in its tree inference procedure.

By way of explanation, we assume that the syntactic trees we compare against (given by the Stanford parser) become more right-branching if punctuation is removed. A simple example is the period at the end of the sentence: this is always attached to a high-level constituent in the correct tree (often to Root), while right-branching attaches it to the most deeply embedded constituent. So this period is always incorrectly predicted by the right-branching baseline, if punctuation is left in.

To further elucidate this issue, we also compute the agreement of various models with a right-branching baseline. 
In the setting without punctuation, the PRPN sets an initial policy that agrees fairly well with right-branching, and this right-branching bias is reinforced by imitation learning and policy refinement.
However, in the setting with punctuation, the agreement with right-branching changes in the opposite way.
We conjecture that right-branching is a reason why our imitation learning achieves a larger improvement without punctuation. 
Right-branching provides a relatively flat local optimum so that imitation learning can do further exploring with a low risk of moving out of it.

\paragraph{Performance across constituent types.} 
We break down the performance of latent tree induction across constituent types in the setting of keeping punctuation. 
We see that, among the six most common ones, our imitation approach outperforms the PRPN on four types. 
However, we also notice that for the most frequent type (NP), our approach is worse than the PRPN. 
This shows that the strengths of the two approaches complement each other, and in future work ensemble methods could be employed to combine them.

\begin{table}
	\centering
	\resizebox{1.0\linewidth}{!}{
	\begin{tabular}{l r| c  c  c}
		\noalign{\hrule height 1.0pt}
		\multirow{2}{*}{Type}  & \multirow{2}{*}{\# Occur}          & \multirow{2}{*}{ST-Gumbel}    & \multirow{2}{*}{PRPN}  & Imitation \\
		& & & &  (SbS + refine)\\
		\noalign{\hrule height 1.0pt}
		NP    & 69k         & 22.6      & \textbf{53.2}  & 49.5 \\
		VP    & 58k         & ~~4.9       & 49.4  & \textbf{57.0} \\
		S     & 42k         & 44.3      & 63.9  & \textbf{66.0} \\
		PP    & 29k         & 13.9      & \textbf{55.4}  & 52.4 \\
		SBAR   & 12k        & ~~6.9       & 38.9  & \textbf{41.4} \\
		ADJP    & 4k       & 10.6      & 44.2  & \textbf{46.5} \\
		\noalign{\hrule height 1.0pt}
	\end{tabular}
	}
	\caption{Parsing accuracy for six phrase types which occur more
		than 2k times in the MultiNLI development set with keeping punctuation.}
	\label{tab:acc on tags}
\end{table}

\paragraph{Discussion.} 
Our results show the usefulness of a downstream task for unsupervised parsing. Specifically, policy refinement with a semantically oriented objective improves parsing performance by two $F$-score points,  outperforming the previous state-of-the-art PRPN model. This provides evidence against previous studies which have claimed that an external, non-syntactic task such as NLI does not improve parsing performance \cite{isitsyntax, replication}. 
At the same time, our results are compatible with findings of \newcite{hao} that a range of different tree structures yield similar classification accuracy in NLI: we find that the mean NLI accuracy of the ST-Gumbel-only model and our imitation learning model with policy refinement is 69.9\% and 69.2\%, respectively, on the MultiNLI development set. NLI performance seems to be largely unaffected by the syntactic properties of the induced trees.

\footnotetext{$F$-score is not normally distributed. It is therefore appropriate to use the non-parametric bootstrap test.}

An interesting question is why ST-Gumbel improves unsupervised parsing when trained with an NLI objective. It has been argued that NLI as currently formulated is not a difficult task \citep{hypothesis}; this is presumably why models can perform well across a range of different tree structures, only some of which are syntactically plausible. However, this does not imply that the Tree-LSTM will learn nothing when trained with NLI. We can think of its error surface being very rugged with many local optima; the syntactically correct tree corresponds to one of them. If the model is initialized in a meaningful catchment basin, NLI training is more likely to recover that tree.
The intuition also explains why the Tree-LSTM alone achieves low parsing performance and low self-agreement. On a very rugged high-dimensional error surface, the chance of getting into a particular local optimum (corresponding to a syntactically correct tree) is low, especially in RL and ST-Gumbel, which are doubly stochastic. 

We show examples of generated trees in Appendix~\ref{app:samples}.

\section{Conclusion}
We proposed a novel imitation learning approach to unsupervised parsing. We start from the differentiable PRPN model and transfer its knowledge to a Tree-LSTM by step-by-step imitation learning. The Tree-LSTM's policy is then refined towards a semantic objective. We achieve a new state-of-the-art result of unsupervised parsing on the NLI dataset. 
In future work, we would like to combine more potential parsers---including chart-style parsing and shift-reduce parsing---and transfer knowledge from one to another in a co-training setting.



\section*{Acknowledgments}
We would like to thank Yikang Shen and Zhouhan Lin at MILA for fruitful discussions.
FK was supported by the Leverhulme Trust through International Academic Fellowship IAF-2017-019.

\bibliography{acl2019}
\bibliographystyle{acl_natbib}

\appendix
\section{Details of the PRPN}
\label{sec:prpn detail}

\begin{figure}[!b]
	\includegraphics[width=.9\linewidth]{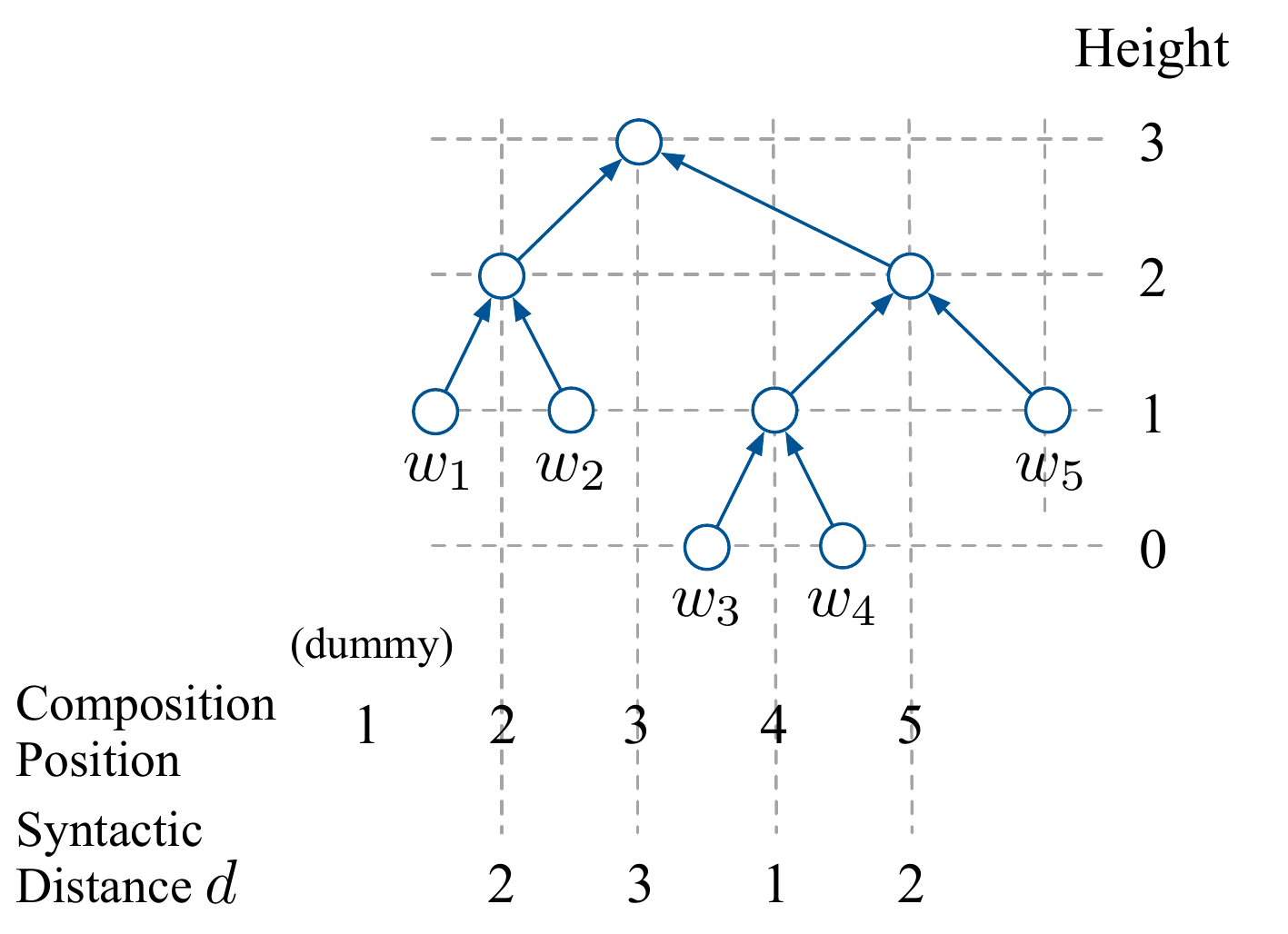}
	\caption{A parse tree with syntactic distance values.}
	\label{fig:tree}
\end{figure}

\begin{figure}[!b]
	\centering
	\includegraphics[width=.90\linewidth]{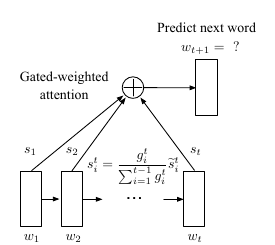}
	\caption{The prediction of the next word in the PRPN language model.}
	\label{fig:predict}
\end{figure}

We now describe in more detail the parsing-reading-predict network (PRPN), proposed by \newcite{prpn}. The PRPN introduces a concept called the \textit{syntactic distance}, illustrated in Figure~\ref{fig:tree}. The syntactic distance $d_t$ is defined as the height of the common ancestor of $w_{t-1}$ and $w_t$ in a tree.

The PRPN uses a two-layer multilayer perceptron (MLP) to estimate $d_t$. The input is the embeddings of the current word and its left context $\bm w_{t-L},\bm w_{t-L+1}, \cdots, \bm w_t$. The output is given by:
\begin{equation}
\widehat d_t = \operatorname{MLP}(\bm w_{t-L}, \bm w_{t-L+1}, \cdots,\bm w_t)
\end{equation}
In fact, absolute distance values are not required, it is sufficient to preserve their order. In other words, if $d_i<d_j$, then it is desired that $\widehat d_i < \widehat d_j$. However, even the order of $d_t$ is not available at training time, and $\widehat d_t$ is learned end-to-end in an unsupervised manner.

The PRPN computes the difference between the distance $d_t$ at the current step and all previous steps $d_j$ for $2\le j<t$. The difference is normalized to the range $[0,1]$:
\begin{equation} 
\alpha_j^t=\dfrac{\operatorname{hardtanh}(\tau(\widehat d_t-\widehat d_j))+1}2
\end{equation}
where $\tau$ is the temperature. 

Finally, a soft gate is computed right-to-left in a multiplicatively cumulative fashion:
\begin{equation}
\label{eqn:gate}
g_i^t=\prod_{j=i+1}^{t-1}\alpha_j^t
\end{equation}
for $1\le i\le t-1$. The gates $g_i^t$ are used to reweight another inner-sentence attention $\widetilde s_i^t$, which is computed as:
\begin{equation}
\widetilde s_i^t=\operatorname{softmax}\{\bm h_i^\top(W[\bm h_{t-1}; \bm w_t])\}
\end{equation}
The reweighed inner-sentence attention $s_i$ then becomes:
\begin{equation}
s_i^t=\dfrac{g_i^t}{\sum_{i=1}^{t-1}g_i^t}\widetilde s_i^t
\end{equation}
and is used to compute the convex combination of attention candidate vectors, which are incorporated in a recurrent neural network to predict the next word, shown in Figure~\ref{fig:predict}. 

\begin{figure*}
\centering
	\includegraphics[width=1.0\linewidth]{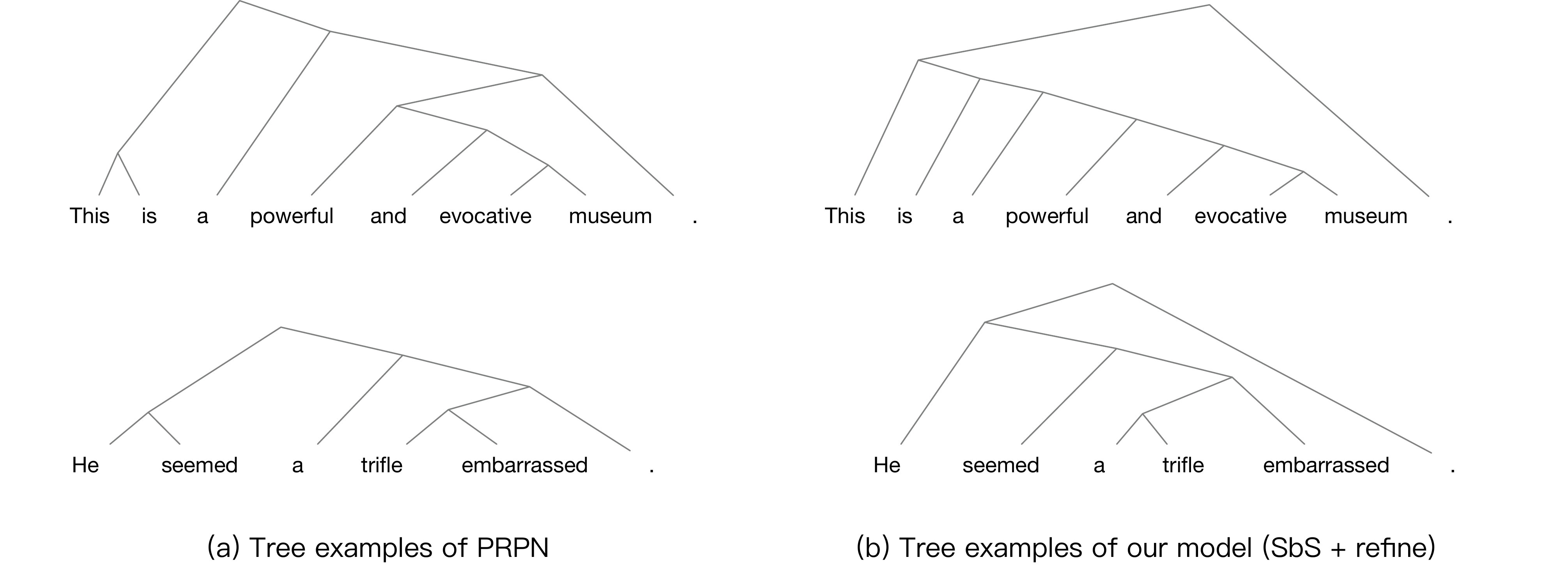}
	\caption{Parse tree examples produced by the PRPN and our model (SbS + refine).}
	\label{fig:tree_samples}
\end{figure*}

\section{Details of Gumbel-Softmax}
\label{sec:gumbel detail}

Gumbel-Softmax can be thought of as a relaxed version of reinforcement learning. It is used in the training of the Tree-LSTM model~\cite{choi}, as well as policy refinement in our imitation learning.
In particular, we use the straight-through Gumbel-Softmax~\cite[ST-Gumbel,][]{st-gumbel}.

In the forward propagation of ST-Gumbel training, the model samples an action---in the Tree-LSTM model, the position of composition---from the distribution $\bm p$ by the Gumbel trick. 
The sampled action can be represented as a one-hot vector $\bm a$, whose elements take the form:
\begin{equation}
\label{eqn:gumbel}
a_i=\left\{
\begin{aligned}
1, &\quad\text{if } i=\operatorname{argmax}_j\{\log(p_j)+g_j\}\\
0, &\quad\text{otherwise}
\end{aligned}
\right.
\end{equation}
where $g_i$ is called the \textit{Gumbel noise}, given by:
\begin{align}
g_i&=-\log(-\log(u_i))\\
u_i&\sim\operatorname{Uniform}(0,1)
\end{align}
It can be shown that $\bm a$ is an unbiased sample from the original distribution $\bm p$~\cite{st-gumbel}.

During backpropagation, ST-Gumbel substitutes the selected one-hot action $\bm a$ given by argmax in Equation~(\ref{eqn:gumbel}) with a softmax operation.
\begin{equation}
\widetilde p_i = \frac{\exp\{(\log(p_i)+g_i)/\gamma\}}{\sum_j\exp\{(\log(p_j)+g_j)/\gamma\}}
\end{equation}
where $\gamma$ is a temperature parameter that can also be learned by backpropagation.

The Tree-LSTM model is trained using the loss in a downstream task (for example, cross-entropy loss for classification problems). Compared with reinforcement learning, the ST-Gumbel trick allows more information to be propagated back to the bottom of the Tree-LSTM in addition to the selected actions, although it does not follow exact gradient computation. For prediction (testing), the model selects the most probable composition according to its predicted probabilities.

\section{Experimental Setup}
\label{setting}
We conduct experiments on the AllNLI dataset, the concatenation of the Stanford Natural Language Inference Corpus \citep{snli} and the Multi-Genre NLI Corpus (MultiNLI; \citealt{multinli}).
As the MultiNLI test set is not publicly available, we follow previous work \citep{isitsyntax,replication} and use the development set for testing. 
For early stopping, we remove 10k random sentence pairs from the AllNLI training set to form a validation set.
Thus, our AllNLI dataset contains 931k, 10k, and 10k sample pairs for training, validation, and test, respectively.

We build the PRPN model and the Tree-LSTM parser following the hyperparameters in previous work \citep{prpn,choi}.\footnotemark\ 
For the SbS training stage, we set $\lambda$ to be 0.03. For the policy refinement stage, the initial temperature is manually set to 0.5.
The PRPN is trained by a language modeling loss on the AllNLI training sentences, whereas the Tree-LSTM model is trained by a cross-entropy loss for AllNLI classification. 

We adopt the standard metric and compute the unlabeled $F$-score of the constituents predicted by our parsing model against those given by the Stanford PCFG Parser (version 3.5.2).
Although the Stanford parser itself may make parsing errors, it achieves generally high performance and is a reasonable approximation of correct parse trees.

\footnotetext{The code bases of the PRPN and the Gumbel Tree-LSTM are available at \url{https://github.com/yikangshen/PRPN} and \url{https://github.com/nyu-mll/spinn/tree/is-it-syntax-release}}

\section{Parse Tree Examples}
\label{app:samples}
In Figure~\ref{fig:tree_samples}, we present a few examples of parse trees generated by the PRPN and by our model (SbS + refine). 

As can be seen, our model is able to handle the period correctly in these examples. Although this could be specified by hand-written rules \citep{diora}, it is in fact learned by our approach in an unsupervised manner, since punctuation marks are treated as tokens just like other words, and our training signal gives no clue regarding how punctuation marks should be processed.

Moreover, our model is able to parse the verb phrases more accurately than the PRPN, including \textit{is a powerful and evocative museum} and \textit{seemed a trifle embarrassed}. This is also evidenced by quantitative results in Table~\ref{tab:acc on tags}.

\newpage

\end{document}